\documentclass{article} % For LaTeX2e
\usepackage{iclr2022_conference,times}
\usepackage{amsmath,amsfonts,bm}

\def\eqref#1{equation~\ref{#1}}
\def\1{\bm{1}}

\DeclareMathAlphabet{\mathsfit}{\encodingdefault}{\sfdefault}{m}{sl}
\SetMathAlphabet{\mathsfit}{bold}{\encodingdefault}{\sfdefault}{bx}{n}

\usepackage{hyperref}
\usepackage{url}
\usepackage{graphicx}
\usepackage{placeins}
\usepackage{float}
\usepackage{adjustbox}

\title{CVCHESS: A DEEP LEARNING FRAMEWORK
FOR CONVERTING CHESSBOARD IMAGES TO
FORSYTH–EDWARDS NOTATION}

\author{Luthira Abeykoon  \\
\texttt{luthira.abeykoon@mail.utoronto.ca} \\
\And
Darshan Kasundra  \\
\texttt{darshan.kasundra@mail.utoronto.ca} \\
\AND
Gawtham Senthilvelan  \\
\texttt{gawthaman.senthilvelan@mail.utoronto.ca} \\
\And
Ved Patel \\
\texttt{ved.patel@mail.utoronto.ca} \\
\AND
}

\iclrfinalcopy 
\begin{document}

\maketitle

\begin{abstract}
Chess has experienced a large increase in viewership since the pandemic, driven largely by the accessibility of online learning platforms. However, no equivalent assistance exists for physical chess games, creating a divide between analog and digital chess experiences. This paper presents CVChess, a deep learning framework for converting chessboard images to Forsyth-Edwards Notation (FEN), which is later input into online chess engines to provide you with the best next move. 

Our approach employs a convolutional neural network (CNN) with residual layers to perform piece recognition from smartphone camera images. The system processes RGB images of a physical chess board through a multistep process: image preprocessing using the Hough Line Transform for edge detection, projective transform to achieve a top-down board alignment, segmentation into 64 individual squares, and piece classification into 13 classes (6 unique white pieces, 6 unique black pieces and an empty square) using the residual CNN. Residual connections help retain low-level visual features while enabling deeper feature extraction, improving accuracy and stability during training. We train and evaluate our model using the Chess Recognition Dataset (ChessReD), containing 10,800 annotated smartphone images captured under diverse lighting conditions and angles. The resulting classifications are encoded as an FEN string, which can be fed into a chess engine to generate the most optimal move.
%######## APS360: Do not change the next line. This shows your Main body page count.
----Total Pages: \pageref{last_page}
\end{abstract}

\section{Introduction}

Chess has seen a dramatic resurgence in popularity, with tournament viewership rising by 55\% from 2020 to 2024 \citep{chesswatch2024}. A key factor behind this surge can be attributed to the growing presence of chess in live streaming, which has made learning the game more accessible than ever. People are interested in chess, and more importantly, people want to get better at chess. While online platforms provide learning methods such as the ‘hint’ button on Chess.com, which recommends the optimal move at a given game state, no such convenience exists for physical chess games.

Our team proposes an easy-to-use tool which can quickly inform users of the most optimal move to make. By using neural networks to detect piece positions from images of physical chess boards at random game states, we can generate a Forsyth-Edwards Notation (FEN) string, a standardized representation of a chessboard. This string can easily be fed into a chess engine to compute the most optimal move. Figure 1 outlines this process. Our system acts as a real-world ‘hint’ button, providing guidance for players of all skill levels. Deep learning works well for this project due to the object-detection abilities of neural networks, particularly in complex spatial settings such as a physical chessboard.

\begin{figure}[H]
\centering
\begin{adjustbox}{center}
\includegraphics[width=0.80\textwidth]{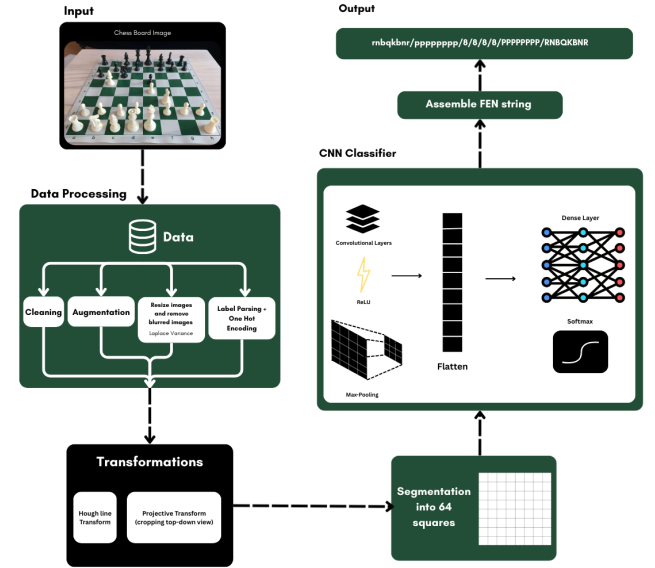}
\end{adjustbox}
\caption{Overview of our model pipeline for real-world chessboard recognition.}
\end{figure}

\section{Background and Related Work}

Much research has been conducted on detecting chess pieces on physical chess boards and keeps on improving as we speak. These studies explore different methods of detecting the board such as using CNNs, x, y gradients, and much more. This section will explore key research in computer vision and its application to identifying chess boards. 

\subsection{End-to-End Chess Recognition \citep{masouris2023}}

Masouris and van Gemert’s work in End-to-End Chess Recognition utilizes the ResNeXt-101 CNN to process the raw images. Their approach skips traditional neural networks and rather trains the model to implicitly learn where the chess board is, understand the spatial layout of the 64 squares and classify what’s in each square (12 unique pieces - 6 black \& 6 white plus empty). This classification is then converted to FEN notation. They correctly classified 15.26\% of ChessReD’s test images which is 7x better than any other current state-of-the-art.

\subsection{Efficient Chess Vision – A Computer Vision Application \citep{wu2022}}

Wu's approach in Efficient Chess Vision – A Computer Vision Application avoids CNNs until after board identification. His method uses a gradient-based technique where the input image is converted to greyscale and x, y gradients are calculated to identify chess board boundaries. Each square within the board undergoes separate identification and pieces are classified into 13 distinct classes before conversion to FEN format. The model achieved 99.5\% accuracy using a medium CNN. 

\subsection{Chessboard and Chess Piece Recognition With the Support of Neural Networks \citep{stojanovic2017}}

The approach in this paper is similar to those discussed above as it employs CNNs to process input images. The chess board identification follows three-steps: detecting straight lines, locating line intersections, and clustering lattice points to identify the chess board. Their piece recognition relies on color and shape descriptors, along with physical properties such as height and area for classification. The results are then converted to FEN format. This model achieved 95\% accuracy in piece detection.

\subsection{Chess-CV \citep{chesscvgithub}}

The method in Chess-CV (a GitHub baked project) projects the chess board onto a 2D plane and employs edge detection algorithms to identify the board. The board is divided into 64 individual squares, where each square is fed into a trained CNN for piece classification. The pieces are classified into 13 distinct classes, and this information is stored in a NumPy array of FEN strings representing the board position. The model achieved 88.9\% accuracy on validation data.

\subsection{ChessVision: Chess Board and Piece Recognition \citep{chessvision2016}}

The approach in ChessVision: Chess Board and Piece Recognition avoids neural networks entirely, and instead utilizes computer vision and machine learning techniques such as SVM and Scale-Invariant Feature Transforms. Their method requires manual selection of the chess board's four corners. For piece recognition, they use variable-sized bounding boxes tailored to different pieces and use SVM classifiers to individually analyze each of the 64 squares. Their output is a visual representation of the board instead of a FEN string. This model achieved 95\% detection accuracy and 85\% classification accuracy.

\section{Data Processing}
\label{headings}

The primary data source used for the project is the Chess Recognition Dataset (ChessReD). This dataset consists of 10,800 images taken from 3 different smartphone cameras of various dimensions with real-world chess games, each at a different board state. The dataset was split in a 60/20/20 ratio, resulting in 6,479 training, 2,192 validation, and 2,129 test images. Each image is annotated with ground truth chess piece positions derived from FEN strings. 

Since the goal is to classify the content of the individual chessboard squares, below is a detailed outline of the steps taken to extract square-level image crops from full-board images. 

\subsection{Board Corner Detection and Perspective Transform}

Each image is first converted to grayscale using cv2.cvtColor(), then smoothed with a Gaussian blur (kernel size 5x5) to reduce noise and enhance edge continuity. Canny edge detection is applied with thresholds of 50x150, which clearly identifies the edges of the chessboard. This is followed by morphological dilation using the same 5x5 kernel to close the gaps in the edge segments due to imperfections such as shadows, chess piece overlaying edges, etc.

Contours are extracted using cv2.findContours() and sorted by area. Only the largest contour with 4 corners and an area exceeding 5\% of the total image is retained as the candidate chessboard. The four corner points are arranged in a consistent orientation, with the white square positioned at the top-left (a8) and bottom-right (h1) corners. This ordered set is then used to compute a projective transformation using cv2.getPerspectiveTransform().

The output of this is a warped 400x400-pixel image representing a top-down-aligned view of the chessboard. This standardization step ensures consistent geometry for downstream processing and model inputs.

\begin{figure}[h]
\begin{center}
\includegraphics[width=0.90\textwidth]{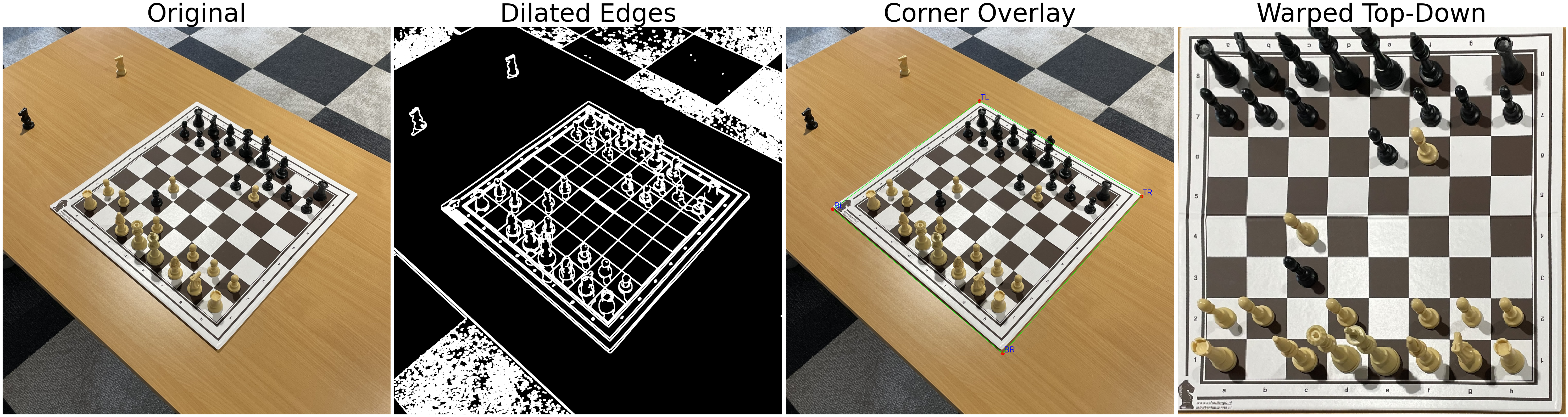}
\end{center}
\caption{Visual Representation of Chess Board Recognition and Perspective Transform}
\end{figure}

\subsection{Data Labeling}

% Once the warped chessboard image is geometrically aligned, we apply the Hough Line Transform to accurately detect the horizontal and vertical grid lines. This step helps ensure that any minor distortions from the perspective transform are corrected, making the square boundaries precise. By locating the intersections of these detected lines, we can reliably segment the board into an exact 8×8 grid.

% The aligned image is then divided into 64 square patches, each corresponding to a single board square. This slicing ensures that every resulting image patch contains only the content of that square, with minimal overlap or misalignment from neighboring squares.

For labeling, the ground truth FEN string representing the full board state is parsed into a 64-element array. Digits in the FEN are expanded into the corresponding number of empty squares, and each piece character is mapped to one of 13 distinct classes: six for white pieces (P, N, B, R, Q, K), six for black pieces (p, n, b, r, q, k), and one for an empty square.

\subsection{New Data Collection and Processing}

For data collection, we recreated the famous 1999 chess match between Garry Lasporov and Veselin Topalov for easy retrieval of FEN strings. The game was played move by move on a physical chessboard with photographs taken at every move, under consistent room lighting. To capture board state variability, each position was photographed from four different angles: a standard top-down view and four additional perspectives around the board as shown in Figure 3. 

\begin{figure}[h]
\begin{center}
\includegraphics[width=0.6\textwidth]{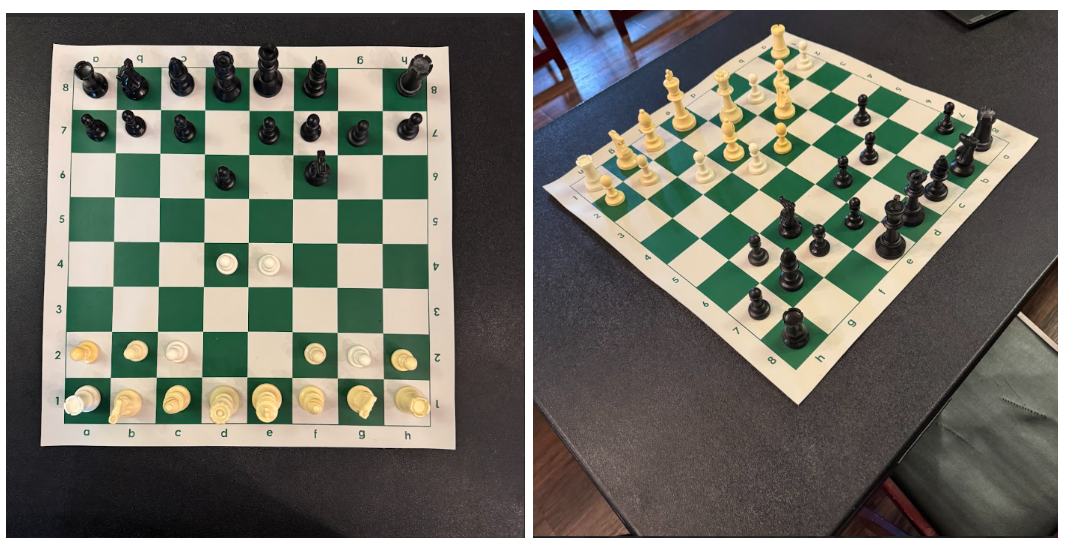}
\end{center}
\caption{Overhead and angle view samples from collected data}
\end{figure}

The primary goal was to create a dataset that pairs each image with an accurate chess position as a FEN string. To do this, we wrote a Python script that does the following: 

\begin{itemize}
    \item Reads the official PGN file of the game and converts each move into its corresponding FEN string using the Python-chess library
    \item Associate each image with the correct FEN label. Since we took five photos per move, the script maps the same FEN to each set of five images.
    \item Stores the image-FEN mappings in a label.json file for later use in training and evaluation. 
  \end{itemize}

In total, the collection process yielded 445 images with 89 unique board states. The resulting dataset provides a multiview representation of each position, which is useful for validating the accuracy of our model and was used to generalize across different camera angles.
\FloatBarrier

\section{Architecture}
\label{gen_inst}

There are several key components the model architecture must address. Firstly, each image is preprocessed as described above, yielding a 400x400 pixel top-down warped image of the chessboard. 

This is then followed by piece recognition where the input is the entire chessboard and the output is the classification of each square. For piece recognition, we took inspiration from a ResNet structure where the CNN has output classes (6 white pieces, 6 black pieces and 1 empty square)\citep{resNet}. The input shape to the CNN is 3x400x400 (RGB image of a single square).

This is fed into the stem layer, which contains a convolutional layer followed by batch normalization, ReLU activation function, and max pooling. The convolutional layer contains a 7x7 kernel that was chosen to capture larger spatial patterns typical in chess pieces. Batch normalization enables more stable training and acts as a regularizer to prevent overfitting. The activation function introduces non-linearity, enabling the model to learn complex patterns during piece recognition, and max pooling reduces computational complexity while preserving spatial information.

Following the above mentioned steps, the image will is fed into three residual layers with pre-activation blocks and skip connections to enable deeper feature learning while avoiding the vanishing gradient problem. Each residual layer progressively increases channel dimensions (64→128→256→512) to extract increasingly complex features, with the first convolution in each block handling channel expansion and the second refining the extracted features. Dropout regularization is applied to prevent overfitting during training. After passing through this process three times, adaptive average pooling reduces spatial dimensions to 8×8, making the tensor of size Bx512x8x8. This is reshaped to (B, 64, 512) to treat each spatial location as one of the 64 chess squares with 512-dimensional feature representations. A final linear classifier maps these features to the 13 piece classes. Once the pieces have been classified, they can easily be converted to FEN notation. 

\begin{figure}[h]
\begin{center}
\includegraphics[width=1.0\textwidth]{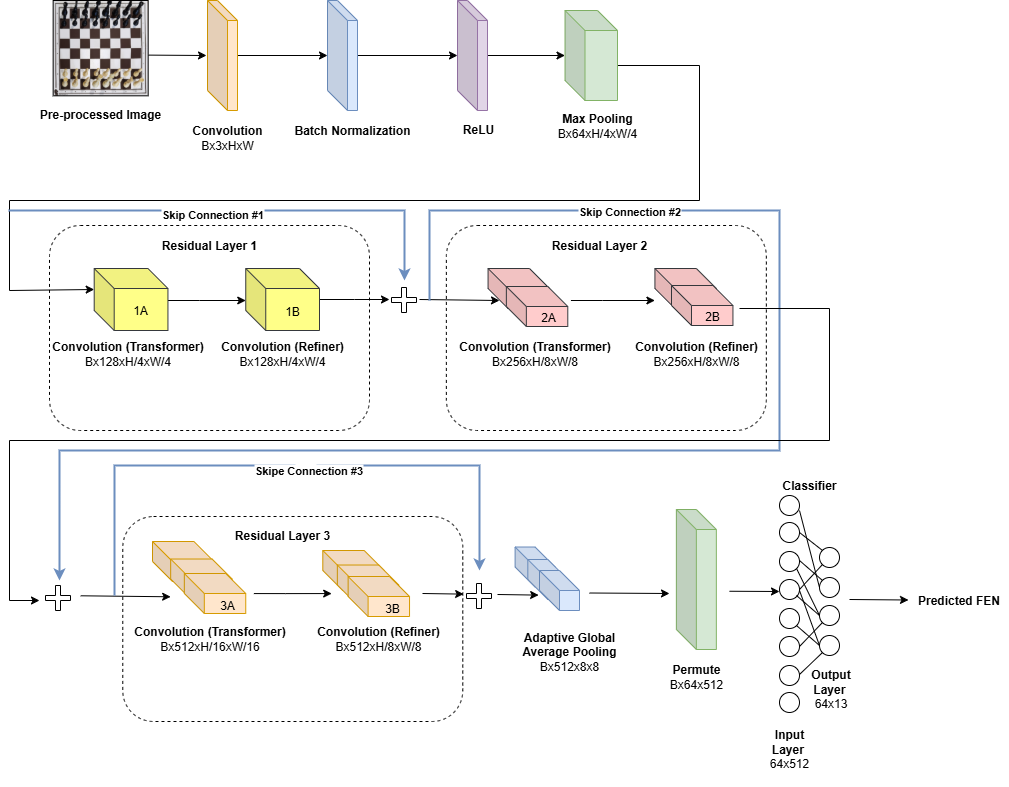}
\end{center}
\caption{Final Model Architecture Overview}
\end{figure}
\FloatBarrier

\section{Baseline Model}
\label{others}

We used a Support Vector Machine (SVM) with an RBF kernel as our baseline, trained on Histogram of Oriented Gradients (HOG) features from individual chessboard squares. To handle the imbalance in the dataset (485 - 37,000 samples per class), we oversampled each class to 938 samples using sklearn.utils.resample.

The pipeline included a StandardScaler for normalization, PCA for dimensionality reduction, and the SVM classifier. Using GridSearchCV with 5-fold stratified cross-validation, we selected pca\_n\_components = 100, svm\_C = 10, and svm\_gamma = 0.01. Data was split 70/30 for training and testing.

The SVM achieved 68\% tile-level accuracy. The normalized confusion matrix shows strong performance on distinctive pieces like queens and bishops (66–71\% accuracy) but much lower accuracy for empty squares (41\%) and black kings (39\%), which were often confused with rooks or pawns. This highlights the model’s strength in recognizing clear shapes and its difficulty with low-contrast or ambiguous tiles.

\begin{figure}[h]
\begin{center}
\includegraphics[width=0.45\textwidth]{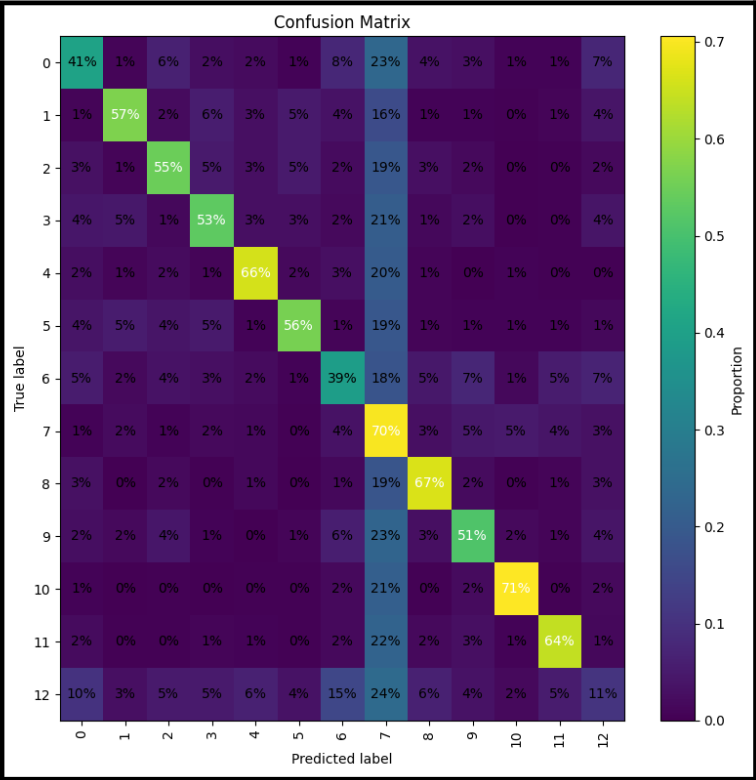}
\end{center}
\caption{Confusion Matrix for Each Piece}
\end{figure}

\section{Quantitative Results}
\label{headings}

Several quantitative evaluations were conducted to assess the model’s accuracy in a comprehensive manner.

The first evaluation involved measuring the number of incorrectly classified pieces on each chessboard, followed by counting how many boards shared the same number of misclassifications. As illustrated in Figure 8, 1145/1790 (63.96\%) boards contained all correctly classified pieces, while the remaining had one or more misclassifications. This represents a significant milestone for the project, as previous methods did not yield such a strong bias toward correctly classified boards.

The second evaluation method involved plotting the training and validation accuracy/loss curves over the course of model training, as shown in Figure 6. These plots provided insight into the general learning trends of the model. The final training accuracy was 99.37\% for all squares and 98.69\% for non-empty squares. In comparison, the test accuracy was 98.93\% for all squares and 97.11\% for non-empty squares. 

Finally, a confusion matrix was generated, as presented in Figure 7. The results indicate that, on average, approximately 95\% of pieces were correctly identified, as reflected by the dominant diagonal entries. The primary misclassifications occurred between visually similar pieces, such as Pawn vs. Bishop (4\%) and Queen vs. King (5\%).

\begin{figure}[h]
\begin{center}
\includegraphics[width=1.0\textwidth]{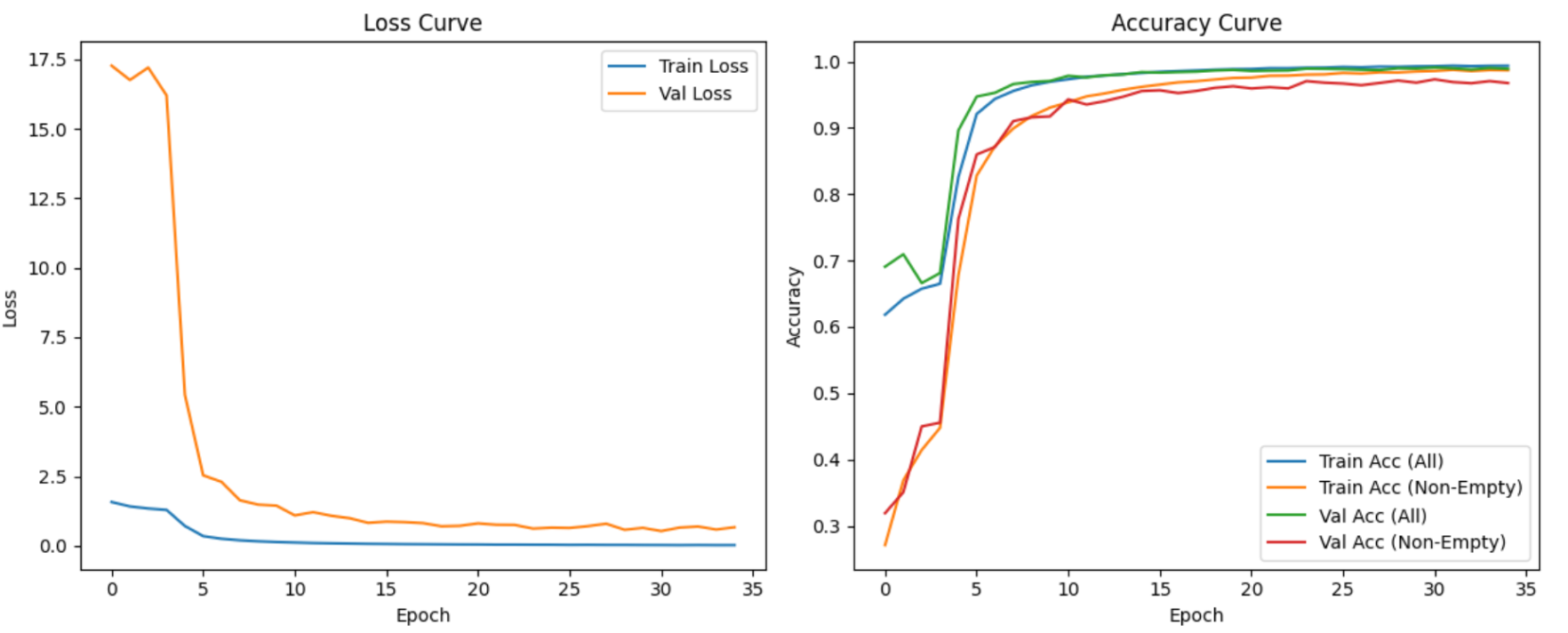}
\end{center}
\caption{Loss / Accuracy Curves on Train and Validation}
\end{figure}

\begin{figure}[h]
\centering
\begin{minipage}{0.48\textwidth}
    \centering
    \includegraphics[width=\linewidth]{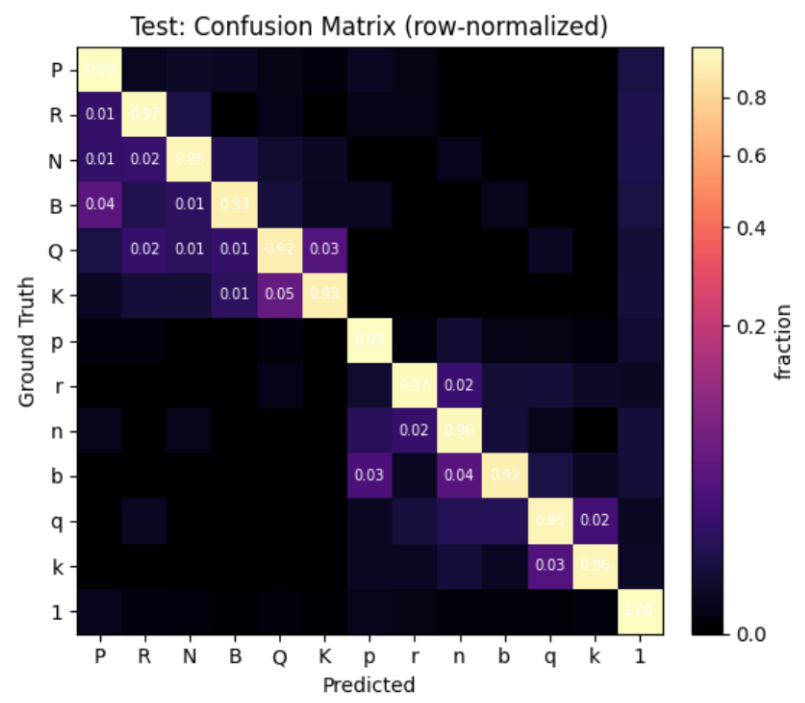}
    \caption{Confusion Matrix for Each Piece on Test Dataset}
\end{minipage}
\hfill
\begin{minipage}{0.48\textwidth}
    \centering
    \includegraphics[width=\linewidth]{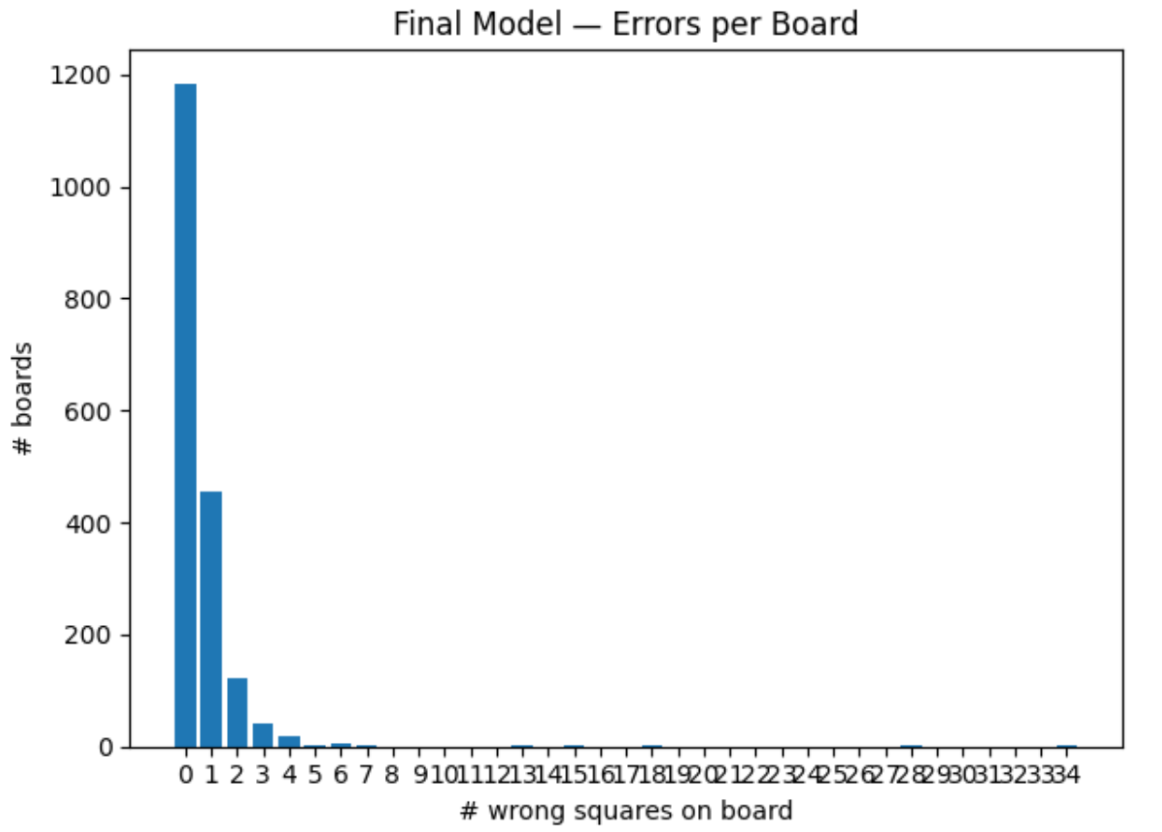}
    \caption{Bar Graph of Correctly Identified Boards on Test Dataset}
\end{minipage}
\end{figure}

\section{Qualitative Results}
\label{others}

The model’s qualitative results highlight both its strengths and weaknesses. For corner detection, predictions on the well-lit from-facing board closely match the ground truth, producing accurate, consistent crops (Figure 9). Slight misalignments appear in images taken at steeper angles, where farside corners tend to shift inward (Figure 10). In terms of FEN string prediction, the model performs well on clear, unobstructed boards, with predicted FEN strings matching the ground truth exactly. However, the performance drops for cases where pieces such as pawns are partially obscured by taller pieces like kings, often leading to square misclassifications, therefore predicting incorrect FEN strings. These errors are more common in batches with warped camera angles or side lighting, where shadows and distortions make classification more challenging.

\begin{figure}[h]
\begin{center}
\includegraphics[width=0.65\textwidth]{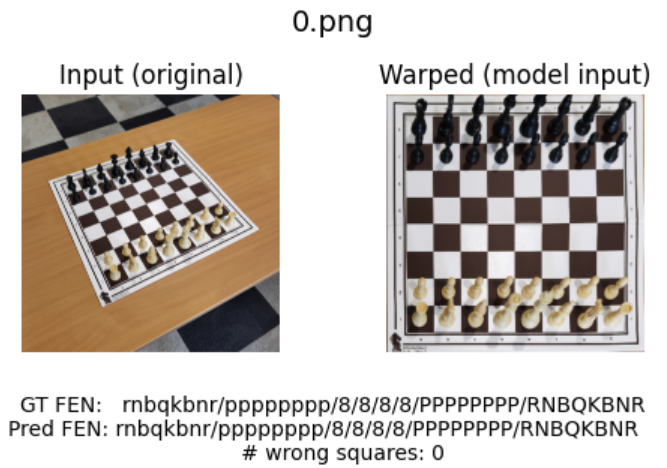}
\end{center}
\caption{Correctly Classified Chessboard}
\end{figure}
\FloatBarrier

\begin{figure}[h]
\begin{center}
\includegraphics[width=0.8\textwidth]{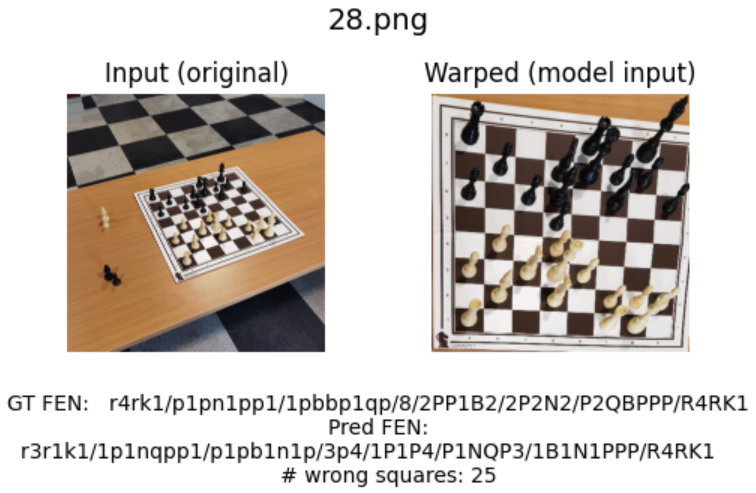}
\end{center}
\caption{Misclassified Chessboard due to slightly steeper angle}
\end{figure}
\FloatBarrier

\section{Evaluation on New Data}

The model has been evaluated on the new dataset made by recreating the famous Kasparov-Topalov 1999 match on an actual chessboard, using an iPhone 15 Pro to collect pictures from varying angles. This introduced all the hardware, resolution, and perspective changes not present in the training set, in addition to some real-world annoyances like glare, shadows, and partial framing of the board. The PGN file of the games was used to obtain the ground truth FEN strings so that the labels perfectly matched.

All preprocessing of images from training, including corner detection, perspective warping, and even a Hough Line Transform for the best possible grid alignment, was kept. In some situations, glare and framing issues would interfere with the detection of the board layout, so an additional edge detection pass attempted to solve the problem. This ensured the pipeline was kept intact, but was able to recover from imperfect real-world inputs.

Across 445 images, the model got 65.17\% per square accuracy, while for the nonempty squares this was 54.06\%. As a result, a full board FEN was generated for for 133/445 total images (29.8\%). The dropoff from the accuracy obtained on the ChessReD testing dataset can be attributed to the physical differences in boards and pieces, occlusion from pawns hidden behind kings, steep angles, bad lighting. All of these factors reduce the effectiveness of the processing pipeline, causing classification to be worsened.

\section{Discussion}

Our final model is a clear step up from previous attempts at working ChessReD. Accuracy and reliability improved significantly once we fixed two main issues: limited training data and a weak preprocessing pipeline. Despite these modifications, our test accuracy remains at approximately 63.9\%, which is still well below perfection; however, it is more than three times higher than the 15.26\% reported in Masouris and van Gemert’s End-to-End Chess Recognition paper \citep{masouris2023}. The main reason for this difference is that our preprocessing pipeline warps images at angles to a top-down view, allowing for a clear view of pieces on individual board tiles (Figure 2).

A notable and somewhat unexpected finding was the substantial impact of preprocessing on overall performance. Without preprocessing, the model failed to generalize to new data; however, even with it, we still lost about one-sixth of our images due to failed board detection. Adding more training data helped just as much as refining preprocessing (during initial stages, we were only using a subsection of ChessReD). The most interesting finding was the trade-off between making the dataset harder (to avoid overfitting) and keeping it clean enough for the model to learn effectively. Background elements like board edges or floor tiles made the task more realistic but also more complex, as it would require a more rigorous edge detection method.

Overall, the model is performing well given the difficulty of the data, the current state of the art, as well as the overall complexity of the project's scope, but it has room to improve. The key lessons were to plan the pipeline early, inspect the dataset closely before training, and understand how preprocessing choices can significantly impact performance.

\section{Next Steps}

With the development of a model that can accurately develop the FEN strings for chess boards with an image taken by an average smartphone, this project presents multiple avenues through which the experience and well-being of chess players can be meaningfully improved. 

One example of which is integration with Stockfish. The FEN strings can be used as input into a Stockfish engine, made accessible by the stockfish python library which can provide the user with the next best move given the current board state \citep{stockFishLibrary}. This integration would resolve the tedious process of manually inputting moves in over-the-board chess games by embedding the system into a mobile app where users can simply take a picture of the board and instantly receive the best move recommendation. 

\section{Ethical Considerations}

The development of this project raises several ethical concerns that should be addressed before any commercial release:

\begin{itemize}
    \item Potential to reduce players’ decision-making skills and game sense in casual play.
    \item Risk of discouraging recreational players if opponents secretly use AI assistance.
    \item Possibility of misuse in competitive settings, undermining fairness and integrity.
    \item Opportunity for professional players to gain unfair advantages through move analysis.
\end{itemize}

Addressing these concerns through clear safeguards and guidelines is essential to prevent cheating and preserve both fair play and the spirit of learning.

\label{last_page}

\bibliography{APS360_ref}
\bibliographystyle{iclr2022_conference}

\end{document}